\documentclass{article}

\usepackage[preprint]{neurips_2023}

\usepackage[utf8]{inputenc} 
\usepackage[T1]{fontenc}    
\usepackage{hyperref}       
\usepackage{url}            
\usepackage{booktabs}       
\usepackage{amsfonts}       
\usepackage{nicefrac}       
\usepackage{microtype}      
\usepackage{xcolor}         
\usepackage{natbib}
\usepackage{graphicx}

\title{Slug Mobile: Test-Bench for RL Testing}

\author{%
  \And
  Jonathan Wellington Morris\\
  Undergrad \\
  UC Santa Cruz\\
  \texttt{jowemorr@ucsc.edu} \\
  \And
  Vishrut Shah\\
  Undergrad \\
  UC Santa Cruz\\
  \texttt{vsshah@ucsc.edu} \\
  \And
  Alex Besanceney\\
  Undergrad \\
  UC Santa Cruz\\
  \texttt{rbesance@ucsc.edu} \\
  \And
  Daksh Shah\\
  Undergrad \\
  UC Santa Cruz\\
  \texttt{dakshah@ucsc.edu} \\
    \And
  Leilani H. Gilpin\\
  Department of Computer Science and Engineering \\
  UC Santa Cruz\\
  \texttt{lgilpin@ucsc.edu} \\
}

\begin{document}

\maketitle

\begin{abstract}
Sim-to real gap in Reinforcement Learning is when a model trained in a simulator does not translate to the real world. This is a problem for Autonomous Vehicles (AVs) as vehicle dynamics can vary from simulation to reality, and also from vehicle to vehicle. Slug Mobile is a one tenth scale autonomous vehicle created to help address the sim-to-real gap for AVs by acting as a test-bench to develop models that can easily scale from one vehicle to another. In addition to traditional sensors found in other one tenth scale AVs, we have also included a Dynamic Vision Sensor so we can train Spiking Neural Networks running on neuromorphic hardware.
\end{abstract}

\section{Introduction}

Autonomous Vehicles (AVs) have been promised to make driving safer for all drivers by replacing human drivers with AVs. \cite{NHTSA} National Highway Traffic Safety Administration (NHTSA) estimates that 94\% of collisions are caused by driver error. \cite{NHTSA2} The main methods for optimal control include classical motion path planning \cite{ferguson2008motion}, end-to-end imitation learning \cite{le2022survey}, and Reinforcement Learning (RL) \cite{aradi2020survey}. Typically, a driving simulator such as CARLA \cite{dosovitskiy2017carla} is used to evaluate vehicle safety and performance \cite{9499331} However, tests in these simulators do not always translate well to real life, resulting in the sim-to-real gap. The sim-to-real gap is a phenomenon where a model trained on a simulator degrades once applied on the real world. \cite{salvato2021crossing} Sim-real-gap can result from a difference in sensor configuration and vehicle dynamics between the simulator and real vehicle. In this paper, we propose a modified version of the F1TENTH vehicle \cite{o2020f1tenth} and MIT Racecar \cite{karaman2017project}, which we call the \textit{Slug Mobile}, which includes updated compute and  sensor array, and a simplified hardware stack. We provide instructions and resources to replicate Slug Mobile here: https://github.com/cruz-control/Slug-Mobile.

\section{Hardware}
\label{hardware}

The vehicle's base frame is the Traxxas Ford Fiesta ST Rally \cite{E-Maxxdude_2017}, the same vehicle specified in \cite{o2020f1tenth}.  The sensors we chose for our car are as follows: RGB Camera, 2D LiDAR, Dynamic Vision Sensor (DVS) also known as Event Cameras \cite{9138762}, and an Inertial Measurement Unit (IMU). We also designed a new mounting solution for the sensors, compute module, and a custom Printed Circuit Board (PCB) which distributes power and signal to the various sensors and the servos.

Most notable change is the inclusion of a DVS camera. We added a DVS camera due to the growing popularity of Spiking Neural Networks (SNN). \cite{10242251} The outputs from DVS cameras are sparse and time-variant, which SNNs can take full advantage to increase efficiency of the driving model.

Another change was a switch from a digital Electronic Speed Controller (ESC) to a PWM ESC. The ESC is then connected to an I2C Servo Driver, PCA9685, which in turn is connected to the NVIDIA Jetson Orin Nano. This was done to simplify the software solution for control. 

The custom PCB has an XT-60 battery connection for a 12V LiPo, which distributes power to each of the sensors. The Jetson Orin Nano's pinout is connected to a pin header on the PCB, enabling sensor communication across the board in addition to power. Signal from the ESC as well as the data and clock signals for I2C communication are routed between the Jetson, IMU, and I2C servo driver. This hardware solution will aid greatly in training and troubleshooting the RL algorithm by reducing complexity of the car.

\begin{figure}
    \centering
    \includegraphics[width=0.5\linewidth]{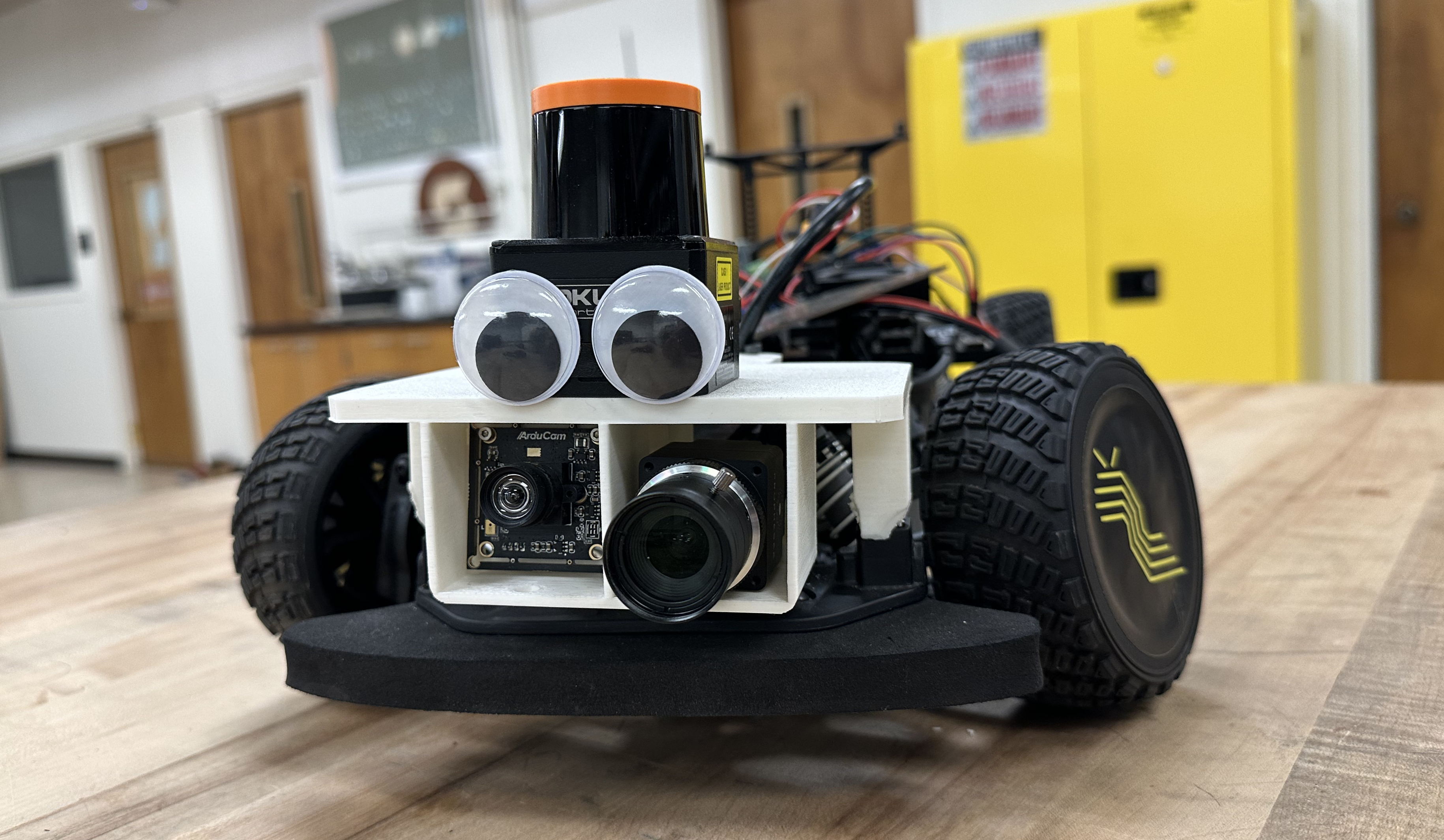}
    \caption{Slug Mobile: One Tenth Autonomous Vehicle for sim-to-real testing}
    \label{fig:car}
\end{figure}

\section{Software}
\label{software}

\begin{figure}
    \centering
    \includegraphics[width=0.5\linewidth]{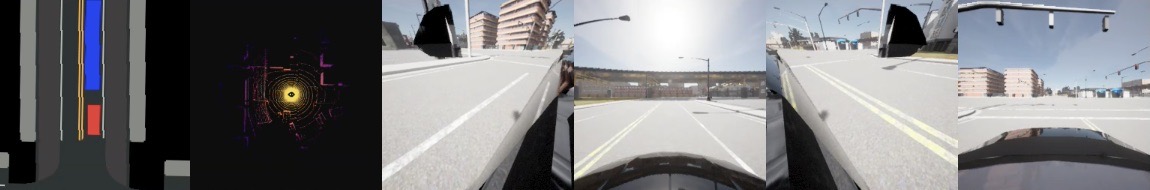}
    \caption{Example visualization of CARLA environment}
    \label{fig:carlagym}
\end{figure}

We deviated from \cite{o2020f1tenth, karaman2017project} by choosing to avoid using ROS and ROS 2. Our software solution for control relies on the NVIDIA jetson-gpio library to communicate with the I2C Servo Driver. This results in code that is easier to read, easier to change, and easier to run. We also have an OpenAI Gym environment, built off of the gym\_carla library \cite{Chen_2020} which works in conjunction with the CARLA simulator. This allows us to simplify running and testing RL algorithms, such as those from Stable Baselines. We customized this environment for compatibility and to train and visualize our results in a virtual environment with a car with a top down LIDAR view and several camera sensors, shown in Figure \ref{fig:carlagym}. We aim to train a model on our gym enviorment and transfer it to our Slug Mobile and evaluate the performance.

\section{Future Works and Conclusion}
\label{future}

One of the roadblocks in Reinforcement Learning for Autonomous Vehicles is the Sim-to-real gap. We aim to use a modified version of the F1Tenth car, called Slug Mobile, to test diffent RL algorithms trained in CARLA to then gauge how well these models transfer from simulation to real vehicles to bridge the sim-to-real gap. We also aim to create a dataset using all the sensors in the car to train models using Imitation Learning, notably with the event camera and neuromorphic hardware to take full advantage of it.

\section{Acknowledgments}
We would like to acknowledge the support of NVIDIA for their support by providing us 2 RTX 4090 GPUs.

\bibliography{bib}

\end{document}